\documentclass[10pt]{article}
\usepackage[explicit]{titlesec}
\usepackage{hyphenat}
\usepackage{ragged2e}
\RaggedRight

\usepackage{ebgaramond}
\usepackage[T1]{fontenc}
\usepackage{amsmath, amsthm}
\usepackage{graphicx}

\usepackage{soul}
\setul{.6pt}{.4pt}

\usepackage{geometry}
\usepackage{fancyhdr}
\usepackage[labelfont={footnotesize,bf} , textfont=footnotesize]{caption}
\makeatletter
\renewcommand\tagform@[1]{\maketag@@@ {\ignorespaces {\footnotesize{\textbf{Equation}}} #1.\unskip \@@italiccorr }}
\makeatother
\titleformat{\section}
  {\normalfont}{\thesection}{1em}{\MakeUppercase{\textbf{#1}}}
\titlespacing\section{0pt}{0pt}{-10pt}
\titleformat{\subsection}
  {\normalfont}{\thesubsection}{1em}{\textit{#1}}
\titlespacing\subsection{0pt}{0pt}{-8pt}

\makeatletter
\newcommand\sixteen{\@setfontsize\sixteen{17pt}{6}}
\renewcommand{\maketitle}{\bgroup\setlength{\parindent}{0pt}
\begin{flushleft}
\sixteen\bfseries \@title
\medskip
\end{flushleft}
\textit{\@author}
\egroup}
\makeatother

\usepackage[sort&compress]{natbib}
\makeatletter
\renewcommand\@biblabel[1]{\textbf{#1.}\hfill}
\makeatother

\bibpunct{}{}{,~}{s}{,}{,}
\title{Human vs. Machine Deception: Distinguishing AI-Generated and Human-Written Fake News Using Ensemble Learning}

\author{
Samuel Jaeger*
$^{a}$
, Calvin Ibenye
$^{b}$
, Aya Vera-Jimenez
$^{ac}$
, Dhrubajyoti Ghosh
$^{a}$\\ \medskip 
$^{a}$School of Data Science and Analytics, Kennesaw State University, Kennesaw, GA \\
$^{b}$Department of Computer Science, Kennesaw State University, Kennesaw, GA \\
$^{c}$Department of Mathematics, Kennesaw State University, Kennesaw, GA\\ \medskip 
Students: sjaeger4@students.kennesaw.edu*, cibenye@students.kennesaw.edu, 
averajim@students.kennesaw.edu \\
Mentor: dghosh3@kennesaw.edu
}

\begin{document}

\vspace*{.01 in}
\maketitle
\vspace{.12 in}

\section*{abstract}
The rapid adoption of large language models has introduced a new class of AI-generated fake news that coexists with traditional human-written misinformation, raising important questions about how these two forms of deceptive content differ and how reliably they can be distinguished. 
This study examines linguistic, structural, and emotional differences between paired human-written misinformation articles and LLM-generated rewrites of those same articles, and evaluates machine learning and ensemble-based methods for distinguishing the two sources of authorship.
A document-level feature representation was constructed using lexical diversity, readability indices, and emotion-based features. Logistic regression, random forest, support vector machines, extreme gradient boosting, a neural network, and an equal-weight probability ensemble were evaluated using repeated article-pair-level train-test splitting so that each human article and its corresponding AI rewrite always remained in the same partition. Across 100 repeated 80:20 splits, the ensemble achieved the highest mean AUC (0.993, SD 0.003) and mean accuracy (0.955, SD 0.014), with mean F1 score 0.955 (SD 0.014). Logistic regression, SVM, random forest, and XGBoost also showed strong performance, while the neural network was somewhat more variable. Readability-based features, particularly the Coleman-Liau index, remained the most informative predictors. These findings indicate that stylistic and structural properties of text provide a robust and interpretable basis for distinguishing human-written misinformation from LLM-generated rewrites of the same underlying misinformation within the controlled paired setting examined here.
\section*{keywords} 
Fake news; AI-generated text; large language models; text classification; readability analysis; lexical diversity; sentiment analysis; ensemble learning; machine learning; misinformation detection
\vspace{.12 in}


\section*{Introduction}

The spread of misinformation and fake news has long posed a serious threat to public discourse, influencing political processes, public health decisions, and societal trust in information systems \cite{lazer2018science, vosoughi2018spread, ghoshthanos}. Traditional forms of misinformation have largely been generated by human authors and often exhibit identifiable linguistic and stylistic characteristics, including irregular sentence structure, variability in readability, and inconsistent emotional expression. These properties have enabled the development of automated detection systems that rely on textual features and machine learning techniques to distinguish deceptive content from legitimate information \cite{shu2017fake, ghosh2026bot}. However, recent advances in artificial intelligence, particularly the emergence of large language models such as GPT-based systems, have fundamentally transformed the nature of misinformation generation. These models are capable of producing highly coherent, context-aware, and stylistically consistent text that closely resembles human writing while minimizing many of the irregularities traditionally associated with deceptive content \cite{brown2020language, openai2023gpt}. Unlike human-written fake news, which is often constrained by cognitive effort and individual writing variability, AI-generated fake news can be produced at scale, rapidly adapted to different topics, and systematically optimized for fluency and readability. This shift significantly amplifies both the volume and the potential impact of misinformation in digital environments. At the same time, it challenges the assumptions underlying existing detection methods, many of which are implicitly designed to capture patterns of human-authored deception. As a result, the emergence of AI-generated fake news raises a critical and timely question: whether machine-generated misinformation differs in measurable and systematic ways from traditional human-written fake news, and whether current detection frameworks remain effective in distinguishing between these two increasingly intertwined forms of deceptive content.

Existing research on fake news detection has primarily focused on distinguishing real news from false or misleading content using machine learning approaches grounded in linguistic, stylistic, and contextual features. Early work relied on handcrafted features such as lexical patterns, syntax, and sentiment \cite{rashkin2017truth, perezrosas2018automatic}, while later studies incorporated user- and network-level information to account for propagation dynamics and social context \cite{shu2019beyond}. More recent advances have leveraged deep learning and transformer-based models to capture higher-order semantic representations, improving performance across diverse datasets \cite{zhou2020survey, oshikawa2020survey}. Related work has also explored stance detection, source credibility, and multimodal signals to enhance detection systems \cite{shu2017fake, zhang2019fake}. Despite these advances, the prevailing framework treats fake news as a single homogeneous category, without distinguishing between different sources of content generation. This assumption becomes problematic in the context of modern generative AI systems, which produce text that is highly coherent, stylistically consistent, and less prone to the irregularities typically exploited by traditional detection methods. While recent studies have examined the credibility and persuasiveness of AI-generated text \cite{clark2021all, kreps2022all}, there remains limited understanding of whether AI-generated fake news differs systematically from human-written fake news at a structural level. Consequently, it is unclear whether models trained on conventional datasets can generalize effectively to machine-generated content, highlighting the need for a more refined understanding of how different forms of misinformation vary according to their origin.

Recent work has begun to examine the more specific problem of distinguishing human-written from LLM-generated text. Guo et al. introduced the Human ChatGPT Comparison Corpus (HC3) and showed that classifier-based systems can distinguish ChatGPT responses from human responses across several domains \cite{guo2023hc3}. Other approaches have focused on model-internal statistical signals rather than supervised classifiers. For example, DetectGPT uses curvature of a language model's log-probability function as a zero-shot signal for machine-generated text and demonstrated strong discrimination on generated news-like text \cite{mitchell2023detectgpt}. At the same time, recent studies have emphasized important limitations of AI-text detection. Sadasivan et al. showed that paraphrasing can substantially reduce the effectiveness of multiple detection strategies, raising questions about robustness as generative systems improve \cite{sadasivan2023reliable}. Liang et al. further demonstrated that several GPT detectors disproportionately misclassified writing by non-native English authors, underscoring the importance of interpretable features and careful domain-specific validation \cite{liang2023bias}. In contrast to approaches based primarily on transformer representations, model likelihoods, or black-box detector scores, the present study focuses on a low-dimensional and interpretable set of stylistic, readability, lexical, and emotional features. It also uses a controlled paired design in which each AI-generated fake-news article is a rewrite of a corresponding human-written fake-news article, allowing the analysis to focus more directly on differences associated with source of generation rather than topic.

Recent work has further shown that the detectability of misinformation may depend on its source of generation. In a controlled comparison using the same feature representation and modeling framework, AI-generated fake news was found to be substantially more separable from real news than human-written fake news, with readability and other interpretable linguistic characteristics accounting for much of this difference \cite{ibenye2026machines}. Related work examining variation across prompting strategies found that interpretable lexical, readability, and emotion-based features retained strong discriminative performance across unseen prompts, suggesting that some linguistic signals may remain stable despite prompt-induced distributional changes \cite{verajimenez2026crossprompt}.
In this study, we investigate whether human-written misinformation can be reliably distinguished from LLM-generated rewrites of the same underlying misinformation and whether these paired texts exhibit systematic differences in their linguistic, structural, and emotional characteristics. To address this question, we construct a document-level feature representation capturing key aspects of writing style, including readability measures, lexical diversity, structural composition, and sentiment and emotion profiles. Using this feature space, we evaluate multiple supervised learning models and develop an ensemble framework \cite{rincy2020,ghosh2025,srivastava2016} to improve classification stability and overall performance. Our results show that human-authored articles and their LLM-generated rewrites can be distinguished with high accuracy under this controlled paired design. Analysis of feature importance indicates that readability and lexical features play a dominant role in this separation, while emotional features contribute comparatively less. In particular, the LLM-generated rewrites exhibit more consistent and uniform stylistic patterns, whereas the human-written articles show greater variability across multiple dimensions. These findings suggest that the rewriting process introduces measurable structural differences that can be effectively captured using computational approaches.


In summary, this work provides a more nuanced perspective on misinformation analysis by moving beyond the traditional real-versus-fake framework and examining differences associated with the source of authorship under a controlled paired rewriting design. The findings show that human-written misinformation and LLM-generated rewrites of the same content are distinguishable and differ in systematic and interpretable ways, particularly in their structural and readability characteristics. These results should not be interpreted as demonstrating universal detectability of AI-generated misinformation; rather, they establish that detectable stylistic differences emerge when an LLM rewrites existing misinformation while preserving its substantive claim and narrative. This distinction is important for understanding how generative systems may alter the linguistic presentation of deceptive content and for guiding future work across broader models, prompts, and generation settings.

\section*{methods and procedures}

\subsection*{Data Description}

The dataset used in this study consists of paired human-written and AI-generated fake news articles constructed under a controlled rewriting framework. Human-written fake news articles were obtained from the Politifact dataset available on Kaggle, which contains fact-checked articles labeled as false or misleading \cite{shu2017fake, shu2019beyond, shu2020fakenewsnet}. For each human-written article, a corresponding AI-generated version was created using GPT-5 through the free version of ChatGPT. 

To generate the AI-based articles, each original human-written fake news article was provided as input to GPT-5 along with a structured prompt instructing it to produce a rewritten version that preserved the same false claims and overall narrative while modifying sentence structure, vocabulary, and organization. The model was instructed to maintain a realistic and coherent news style and to avoid introducing new factual claims. This controlled prompting strategy was intended to make the two members of each pair comparable in underlying content while allowing stylistic and structural characteristics to differ.

The AI-generated articles were produced using GPT-5 through the free version of ChatGPT during November-December 2025. Each original human-written fake-news article was supplied to the model together with the following structured prompt:

\begin{quote}
\small
\textit{You are a professional journalist writing for a research corpus. Create a rewritten version of the ARTICLE below that preserves the SAME FALSE CLAIMS and the same overall storyline, but changes sentence structure, vocabulary, and paragraph order to produce a clearly paraphrased version. Maintain a realistic news style (serious, coherent, non-satirical). Do NOT add new factual claims or citations. Output only the rewritten article text.}
\end{quote}

This prompting strategy was designed to preserve the substantive false claim and overall narrative of each source article while inducing variation primarily in wording, sentence structure, and organization.

\begin{table}[ht]
\centering
\caption{Illustrative example of a paired human-written and AI-generated fake-news article. Short excerpts are shown for readability.}
\label{tab:example_pair}
\resizebox{\textwidth}{!}{%
\begin{tabular}{p{0.16\textwidth} p{0.76\textwidth}}
\hline
\textbf{Source} & \textbf{Excerpt} \\
\hline

Human-written &
``Reuters reported that Nancy Pelosi bought 10 million shares of cannabis stock. On Oct. 6, President Joe Biden said he would pardon all people convicted of simple marijuana possession offenses under federal law and directed Cabinet members to review how the drug is classified. Later that day, a claim started to spread on social media that House Speaker Nancy Pelosi had bought millions of shares of cannabis stock before Biden's announcement.'' \\[6pt]

AI-generated &
``Following President Joe Biden's announcement on Oct. 6 that he would pardon all federal convictions for simple marijuana possession and direct a review of the drug's federal classification, a viral claim surfaced on social media asserting that House Speaker Nancy Pelosi had purchased millions of shares of cannabis stock. An Oct. 7 Instagram post stated, `Breaking: Nancy Pelosi purchased 10,000,000 shares of \$WEED 4 days ago,' attributing the claim to Reuters.'' \\

\hline
\end{tabular}%
}
\end{table}

Table~\ref{tab:example_pair} illustrates the paired design used throughout the study. The AI-generated version preserves the same topic and underlying false claim as the human-written source article while altering wording, sentence structure, and organization.

Each original article and its AI-generated rewrite were assigned a common article-pair identifier. After removal of missing or empty texts, the final dataset consisted of 1000 documents forming 500 complete pairs, with 500 human-written and 500 AI-generated articles. Individual documents served as classification observations, but the paired structure was explicitly preserved during all data splitting and cross-validation procedures. This design enables a direct comparison between human-written and AI-generated fake news while reducing confounding due to topic or underlying claims.

\subsection*{Feature Extraction}

A comprehensive set of document-level features was constructed to capture differences in linguistic style, structural composition, readability, and emotional expression between human-written and AI-generated fake news. These features reflect multiple dimensions of writing that have been widely used in text analysis and deception detection. All features were aggregated at the document level. Structural features were used for descriptive characterization; the classification models used the lexical diversity, readability, and emotion features described below.

First, structural features were computed to characterize overall writing style. These included measures of document length, such as total number of characters and words, as well as sentence-based metrics including estimated sentence count and average sentence length. Additional features captured punctuation usage and capitalization patterns, which are commonly used indicators of stylistic consistency and have been shown to be useful in authorship and deception analysis \cite{stamatatos2009survey}.

Second, lexical diversity was quantified using the type-token ratio, defined as the proportion of unique words relative to the total number of words within a document. This measure reflects vocabulary richness and variability and is widely used in stylometric and linguistic analysis to differentiate writing styles \cite{templin1957certain, tweedie1998measures}.

Third, readability features were included to assess the complexity and structure of the text. Standard readability indices were computed, including the Flesch Reading Ease score \cite{flesch1948new}, Flesch-Kincaid Grade Level \cite{kincaid1975derivation}, SMOG index \cite{mclaughlin1969smog}, and Coleman-Liau index \cite{Coleman1975simple}. These measures capture variation in sentence length, word complexity, and overall readability, and are widely used in evaluating written text. All readability indices were computed using the textstat\_readability function in the quanteda.textstats R package. For the Coleman-Liau index, the expected cloze percentage form, $141.8401 - 0.214590L + 1.079812S$, was used, where $L$ is the number of letters per 100 words and $S$ the number of sentences per 100 words; higher values indicate text that is easier to read.

Finally, emotional and sentiment-related features were derived using the NRC Emotion Lexicon \cite{mohammad2013nrc}, which categorizes words into affective dimensions such as anger, fear, joy, sadness, trust, and anticipation. For each document, the proportion of words associated with each emotion category was computed to reflect relative emotional intensity while controlling for document length. Emotion-based features have been shown to be informative in distinguishing different types of textual content, including deceptive and persuasive writing.


\subsection*{Data Preprocessing}

Out-of-sample performance was evaluated using 100 repeated 80:20 train-test splits performed at the article-pair level. For each repetition, 80\% of the 500 article pairs were assigned to the training set and the remaining 20\% to the test set. Both members of a pair were always assigned to the same partition, preventing the human-written article from appearing in one partition while its corresponding AI-generated rewrite appeared in the other.

All centering and scaling parameters were estimated from the training data only and were then applied unchanged to the corresponding held-out test data. The dataset was balanced across the two classes, with equal numbers of human-written and AI-generated fake news articles, so no additional class weighting or resampling was required. No dimensionality reduction or feature selection was applied because the feature set was deliberately designed to remain interpretable and moderate in size.

\subsection*{Classification Models}

To distinguish between human-written and AI-generated fake news, we evaluated logistic regression, random forest (RF), support vector machines (SVM), extreme gradient boosting (XGBoost), and a feedforward neural network (NN). Logistic regression was used as a linear baseline. Random forest and XGBoost were used to capture nonlinear relationships and interactions, and the SVM used a radial basis function kernel. The neural network consisted of a single hidden layer. In the primary analysis, all models used fixed settings: random forest with 500 trees and the default $m_{\text{try}} = \lfloor\sqrt{15}\rfloor = 3$; a radial-kernel SVM with $C = 1$ and $\gamma = 1/15$; XGBoost with maximum depth 4, learning rate 0.1, 200 boosting rounds, and row and column subsampling of 0.8; and a neural network with 10 hidden units, no weight decay, and at most 200 iterations. All models used the same 15 standardized features: type-token ratio, four readability indices, and ten NRC emotion and sentiment proportions. The use of multiple statistical and machine-learning paradigms also reflects the broader principle that comparing complementary modeling approaches can provide a more robust assessment of predictive structure \cite{ghosh2026efficacy}.

The primary performance analysis used the same model specifications across the 100 repeated pair-level train-test splits. In addition, a nested robustness analysis was conducted using 10 outer pair-level 80:20 splits. Within each outer training set, hyperparameters were selected by grouped five-fold cross-validation, with both documents from each article pair retained within the same fold and AUC used as the tuning criterion. The tuning grids and most frequently selected settings are summarized in Table~\ref{tab:tuning}. XGBoost did not yield a single dominant configuration, indicating that several settings produced similar cross-validated performance. In this nested analysis, the ensemble retained strong performance (mean accuracy 0.962 and mean AUC 0.995), supporting the robustness of the primary results to hyperparameter selection.

\begin{table}[ht]
\centering
\caption{Hyperparameter tuning used in the grouped five-fold cross-validation robustness analysis.}
\label{tab:tuning}
\resizebox{\textwidth}{!}{%
\begin{tabular}{llll}
\hline
Model & Parameter(s) tuned & Candidate values & Most frequently selected setting \\
\hline
Logistic regression & None & - & - \\
Random forest & $mtry$ & 2, 3, 7 & $mtry=7$ (9/10 outer splits) \\
SVM (radial) & $\sigma$, $C$ & $\sigma$: 0.01, 0.05, 0.10; $C$: 0.5, 1, 2, 4 & $\sigma=0.01$, $C=4$ (9/10) \\
XGBoost & rounds, depth, learning rate & rounds: 100, 200; depth: 2, 4; $\eta$: 0.05, 0.10 & No single dominant setting \\
Neural network & hidden units, weight decay & units: 5, 10, 15; decay: 0, 0.001, 0.01 & 15 units, decay 0.01 (9/10) \\
\hline
\end{tabular}%
}
\end{table}

\subsection*{Ensemble Framework}

In addition to individual models, we constructed an ensemble classifier to improve predictive stability and overall performance. The ensemble combines predictions from multiple base learners by averaging their predicted probabilities for each class. Specifically, for a given observation, the final predicted probability was obtained by taking the mean of the predicted probabilities from all models considered in this study.
The motivation for this approach is that different models capture different aspects of the feature space and may vary in their sensitivity to noise or specific feature patterns. By aggregating predictions across models, the ensemble reduces model-specific variance and leads to more robust and stable classification performance.
The final class label was assigned based on the aggregated probability using a standard threshold of $0.5$. The ensemble model was evaluated alongside individual models to assess its effectiveness in improving classification accuracy and consistency.

\subsection*{Evaluation Metrics}

Model performance was evaluated using accuracy, AUC, precision, recall (sensitivity), specificity, and F1 score. AI-generated fake news was treated as the positive class. Accuracy measures the overall proportion of correctly classified observations; precision measures the proportion of predicted AI-generated articles that were correctly classified; recall measures the proportion of AI-generated articles correctly identified; specificity measures the proportion of human-written articles correctly identified; and the F1 score summarizes the balance between precision and recall. AUC provides a threshold-independent measure of discrimination. For the primary analysis, each metric was computed on the held-out test set within each of the 100 pair-level splits and summarized by its mean and standard deviation across repetitions. AUC was also used separately to describe the discriminative ability of individual features.

\subsection*{Feature Importance Analysis}

Feature importance was examined using both model-based and feature-level approaches. For random forest, importance was measured using Mean Decrease Gini. For XGBoost, importance was measured using gain, which quantifies the improvement in the objective attributable to splits involving a feature. For visualization, importance values were scaled separately within each model by dividing each value by the maximum importance value for that model, so that the most important feature had a scaled value of 1.

In addition, the discriminative ability of each feature was examined individually using AUC. This combined approach provides both model-specific and feature-specific perspectives on the variables driving separation between human-written and AI-generated fake news.

\section*{results}

\subsection*{Overall Model Performance}

\begin{figure}[ht]
\centering
\includegraphics[width=\textwidth]{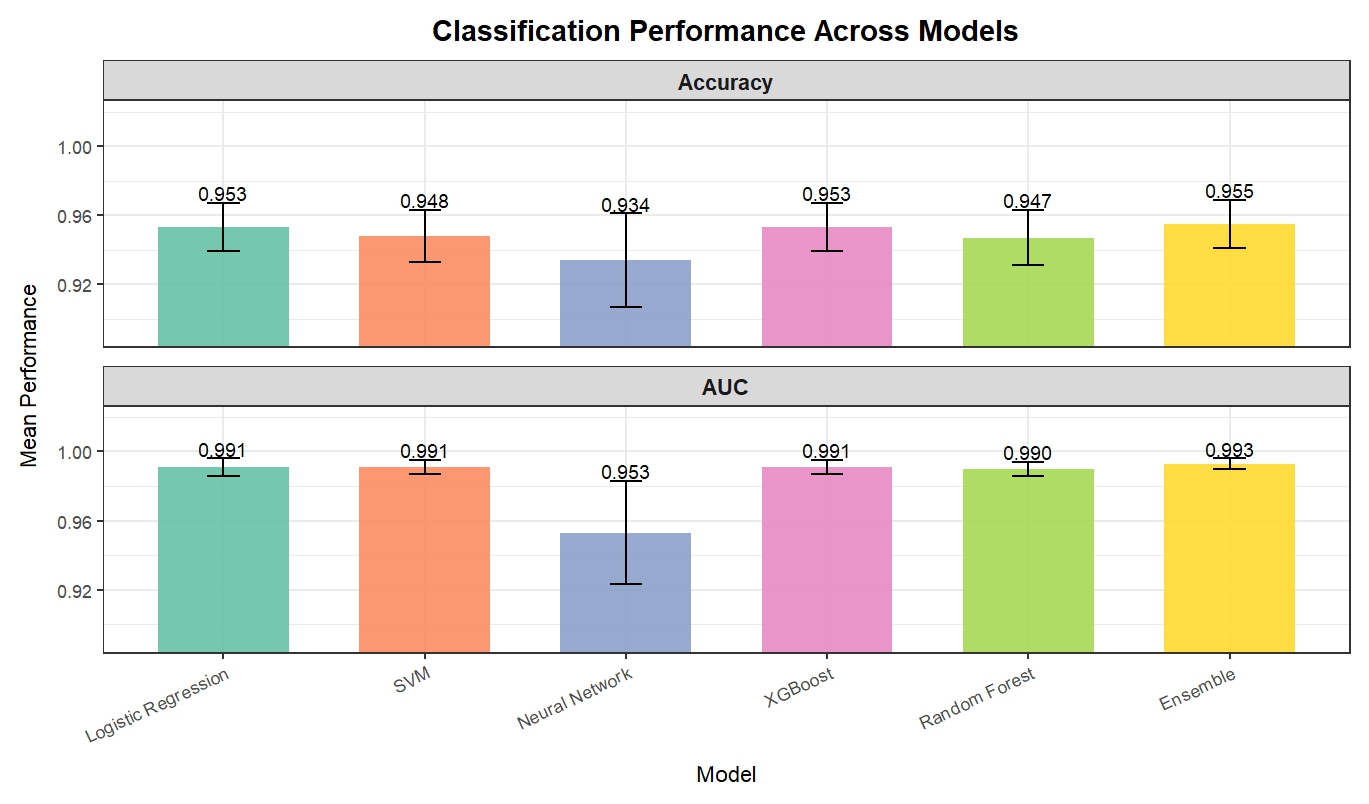}
\caption{Mean classification accuracy and area under the receiver operating characteristic curve (AUC) across 100 repeated 80:20 article-pair-level train-test splits. Error bars represent one standard deviation across repetitions.}
\label{fig:performance}
\end{figure}

Performance across the 100 repeated pair-level 80:20 train-test splits is summarized in Figure~\ref{fig:performance} and Table~\ref{tab:performance}. All models showed strong discrimination, with mean accuracy ranging from 0.934 to 0.955 and mean AUC ranging from 0.953 to 0.993. The ensemble achieved the highest mean AUC (0.993, SD 0.003), mean accuracy (0.955, SD 0.014), and mean F1 score (0.955, SD 0.014). Logistic regression and XGBoost each achieved mean accuracy of 0.953, with mean AUC of 0.991. SVM and random forest also performed strongly, with mean AUC values of 0.991 and 0.990, respectively. The neural network had lower and more variable performance than the other models, with mean accuracy of 0.934 (SD 0.027) and mean AUC of 0.953 (SD 0.030). Overall, the ensemble provided the strongest average performance, although several individual models performed comparably.

\begin{table}[ht]
\centering
\caption{Performance across 100 repeated 80:20 article-pair-level train-test splits. Values are mean (SD).}
\label{tab:performance}
\resizebox{\textwidth}{!}{%
\begin{tabular}{lcccccc}
\hline
Model & Accuracy & AUC & Precision & Recall & Specificity & F1 \\
\hline
Ensemble & 0.955 (0.014) & 0.993 (0.003) & 0.954 (0.020) & 0.956 (0.022) & 0.954 (0.021) & 0.955 (0.014) \\
Logistic & 0.953 (0.014) & 0.991 (0.005) & 0.951 (0.020) & 0.956 (0.022) & 0.951 (0.021) & 0.953 (0.014) \\
NN & 0.934 (0.027) & 0.953 (0.030) & 0.933 (0.032) & 0.935 (0.041) & 0.932 (0.035) & 0.934 (0.029) \\
RF & 0.947 (0.016) & 0.990 (0.004) & 0.957 (0.018) & 0.937 (0.025) & 0.957 (0.019) & 0.947 (0.016) \\
SVM & 0.948 (0.015) & 0.991 (0.004) & 0.944 (0.018) & 0.953 (0.025) & 0.943 (0.020) & 0.948 (0.015) \\
XGBoost & 0.953 (0.014) & 0.991 (0.004) & 0.959 (0.018) & 0.947 (0.022) & 0.959 (0.019) & 0.953 (0.014) \\
\hline
\end{tabular}%
}
\end{table}

The nested grouped cross-validation robustness analysis produced similarly strong results after hyperparameter tuning. Across 10 outer pair-level splits, the ensemble achieved mean accuracy $0.962$ (SD $0.014$) and mean AUC $0.995$ (SD $0.003$), while SVM achieved mean accuracy $0.960$ (SD $0.011$) and mean AUC $0.994$ (SD $0.004$). These findings indicate that the primary conclusions are not sensitive to the choice of a single train-test partition or to the use of fixed model settings. For comparison, a document-level 80:20 split that ignores the pairing yielded ensemble accuracy 0.955 and AUC 0.992; the pair-level estimates are of similar magnitude, indicating that the findings were not driven by leakage between paired documents.

\subsection*{Feature Importance Analysis}

\begin{figure}[ht]
\centering
\includegraphics[width=\textwidth]{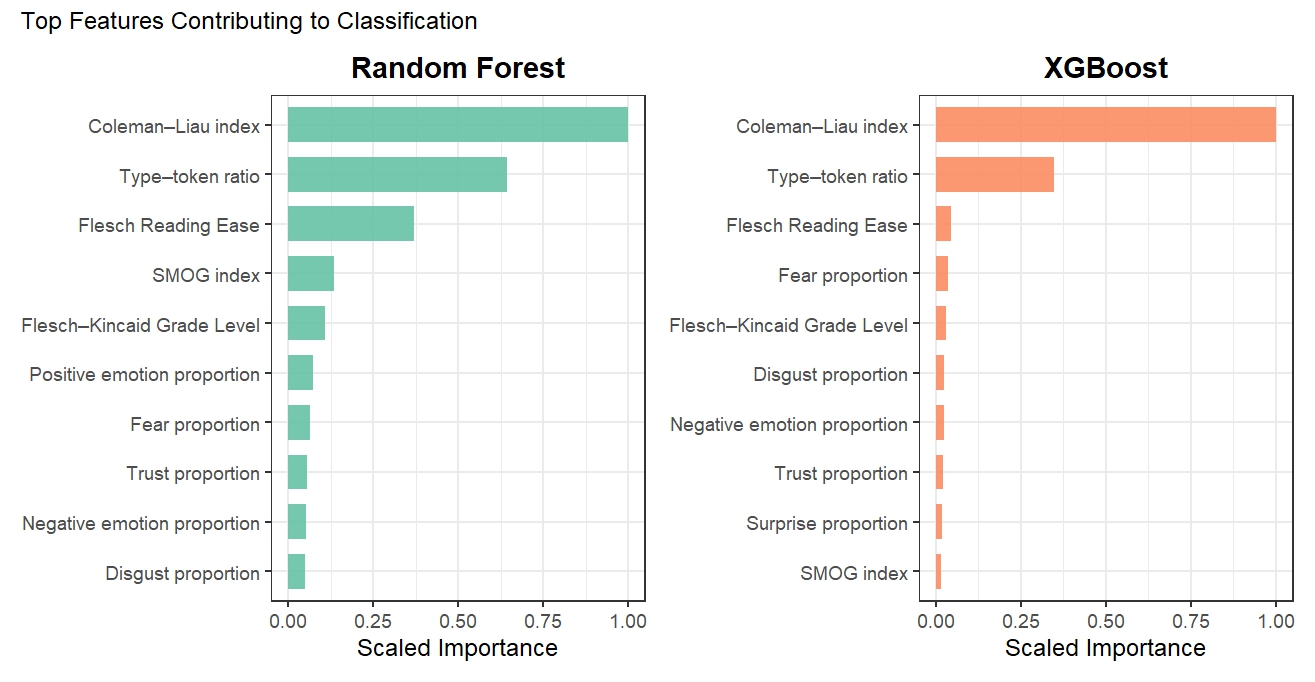}
\caption{Top features contributing to classification for the random forest and XGBoost models. Random forest importance is based on mean decrease in Gini impurity, while XGBoost importance is based on gain. Importance values are scaled within each model so that the largest value equals 1. Readability-related features dominate both rankings, with the Coleman-Liau index contributing most strongly.}
\label{fig:importance}
\end{figure}

Feature importance was assessed using both random forest and XGBoost models. The top contributing features based on scaled importance scores are shown in Figure~\ref{fig:importance}.
It is observed that readability-based features dominate the importance rankings. In particular, the Coleman-Liau index is identified as the most influential feature across both models, followed by type-token ratio and the Flesch readability score. This indicates that structural and stylistic characteristics of the text play a central role in distinguishing between human-written and AI-generated fake news.
Emotion-based features, including proportions of positive, negative, fear, trust, and disgust-related words, exhibit comparatively lower importance. While these features contribute to the model, their individual impact is limited relative to readability metrics. Consistency in feature rankings across both models suggests robustness of these findings.

\subsection*{Discriminative Power of Individual Features}

\begin{figure}[ht]
\centering
\includegraphics[width=\textwidth]{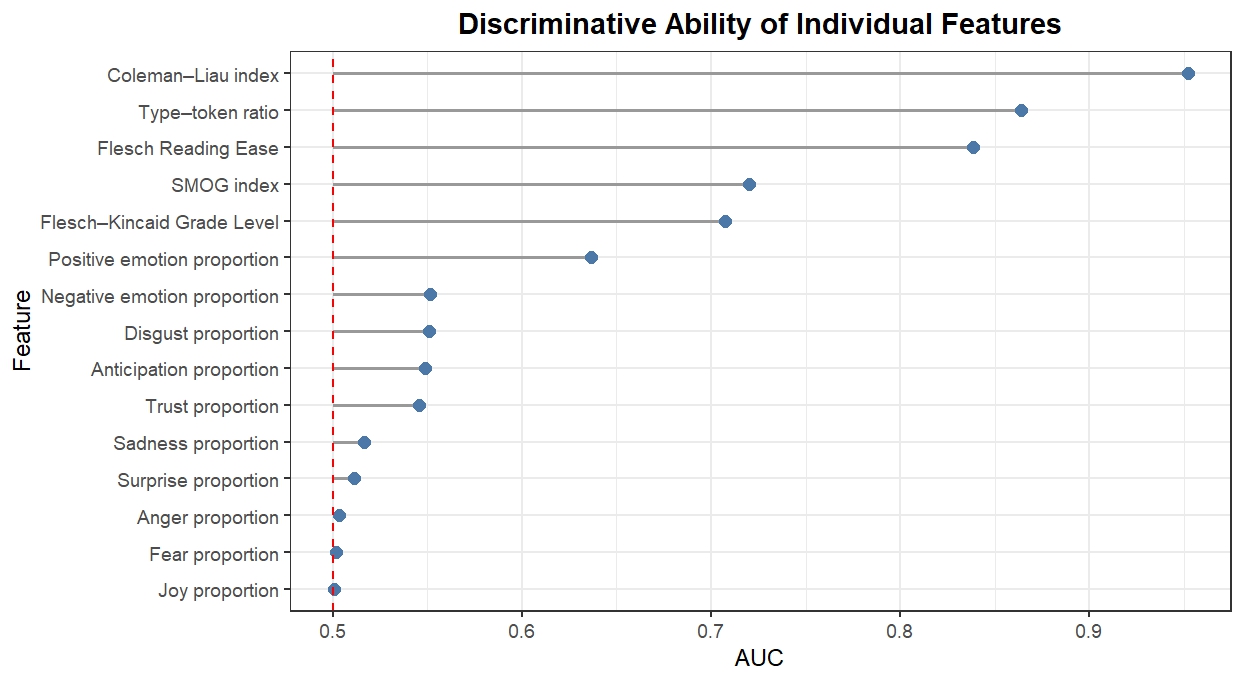}
\caption{Discriminative ability of individual features measured by area under the receiver operating characteristic curve (AUC). Readability-related features show the strongest standalone discrimination, whereas most emotion-based features perform close to chance level.}
\label{fig:single_auc}
\end{figure}

The discriminative ability of individual features was evaluated using AUC. The results are presented in Figure~\ref{fig:single_auc}.
Among all features, the Coleman-Liau index achieves the highest AUC (0.952), followed by type-token ratio (0.864) and the Flesch readability score (0.838). These results indicate that readability-related features possess strong standalone predictive power. In contrast, most emotion-based features exhibit AUC values close to 0.5, suggesting performance comparable to random guessing when used independently.
These findings indicate that while emotional features may provide supplementary information in multivariate models, they are insufficient as primary predictors. In contrast, structural properties of text are shown to be highly informative even in isolation.

\subsection*{Distributional Differences in Readability}

\begin{figure}[ht]
\centering
\includegraphics[width=\textwidth]{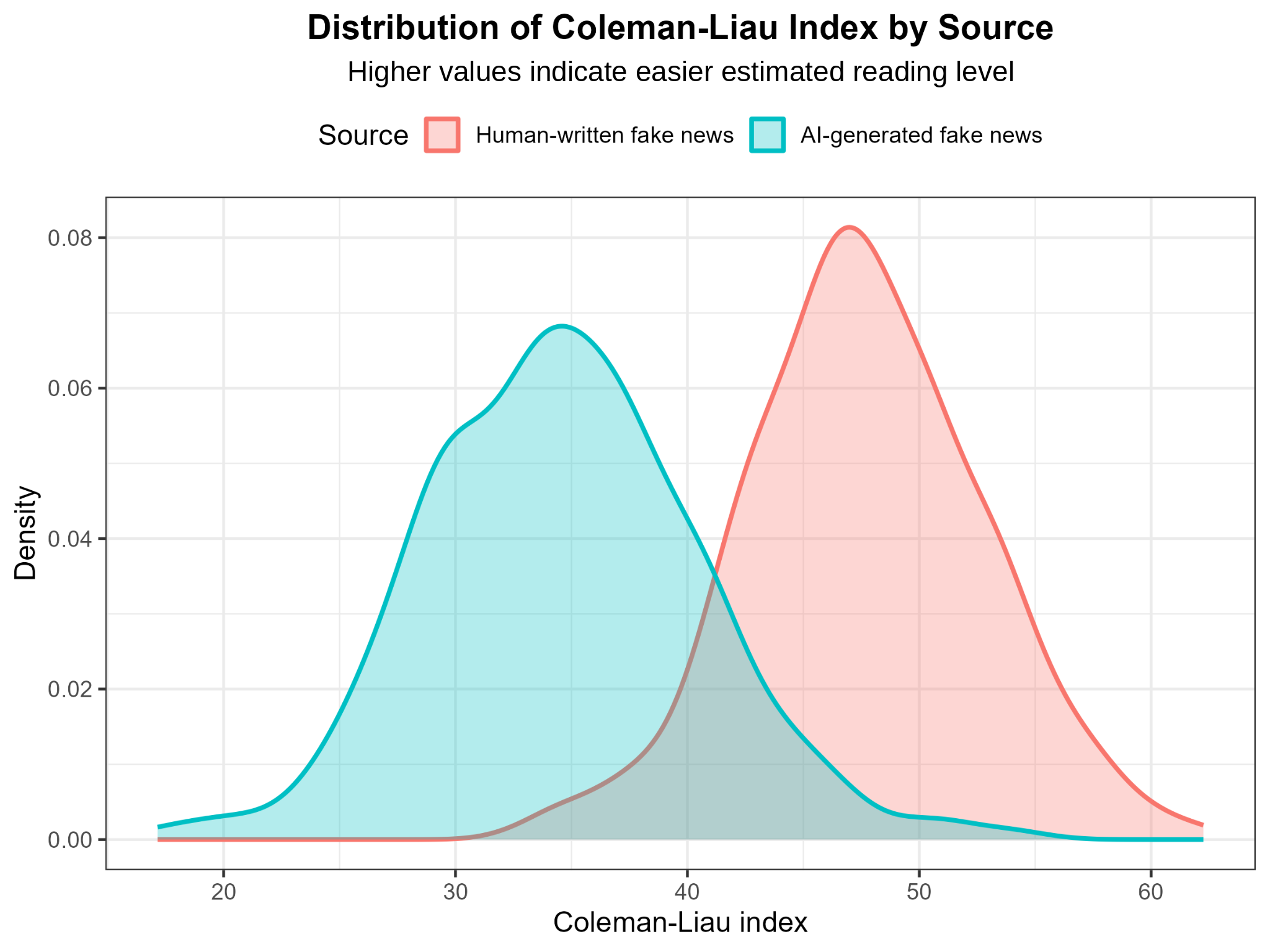}
\caption{Distribution of the Coleman-Liau index (expected cloze percentage form) for human-written and AI-generated fake news. Higher values indicate text estimated to be easier to read. Human-written articles tend to have higher values, whereas AI-generated articles are concentrated at lower values.}
\label{fig:density}
\end{figure}

To further examine the role of readability, the distribution of the Coleman-Liau index across the two classes was analyzed. The corresponding density plots are shown in Figure~\ref{fig:density}.
A clear shift between the two distributions is observed. 
Human-written fake news is associated with higher Coleman-Liau values, indicating text estimated to be easier to read. In contrast, AI-generated fake news is concentrated at lower Coleman-Liau values, indicating greater estimated reading difficulty.
Although some overlap is present, the difference in central tendency is pronounced, providing strong evidence that readability serves as a key distinguishing characteristic. This observation is consistent with the feature importance and individual AUC analyses.

\section*{discussion}

The results demonstrate that distinguishing between human-written and AI-generated fake news can be achieved with consistently high accuracy using relatively simple text-based features. Across the 100 repeated article-pair-level train-test splits, classification performance remained uniformly strong, with mean accuracy exceeding 93\% and mean AUC values ranging from 0.953 to 0.993. This suggests that the differences between the two classes are not subtle artifacts of specific models, but instead reflect systematic and measurable distinctions in the underlying text.
A key finding is the dominant role of readability-based features in driving classification performance. Metrics such as the Coleman-Liau index, type-token ratio, and Flesch readability score were consistently identified as the most informative predictors across models. In particular, the Coleman-Liau index alone achieved a high standalone AUC, indicating that substantial discriminatory information is captured by this feature in isolation. This highlights the importance of structural and stylistic properties of text, suggesting that differences in sentence construction, word complexity, and lexical diversity serve as primary signals for distinguishing between human and AI-generated content.
In contrast, emotion-based features contributed relatively little to classification performance. While proportions of words associated with emotions such as fear, trust, or disgust were included in the models, their individual predictive power was limited, with AUC values close to random guessing. This finding should be interpreted in light of the paired rewriting design. Because each AI-generated article was conditioned on the corresponding human-written misinformation article and instructed to preserve the same false claim and overall narrative, the rewriting process may also have preserved aspects of the source article's emotional framing. Consequently, the limited discriminative ability of emotion-based features may reflect similarity inherited from the original text rather than an absence of emotional content in AI-generated misinformation. Emotional features should therefore not be interpreted as generally uninformative for distinguishing human- and AI-generated text outside the controlled rewriting setting examined here.

The observed distributional differences in readability further reinforce these findings. 
Human-written fake news was associated with higher Coleman-Liau values, corresponding to text estimated to be easier to read, while AI-generated text exhibited lower Coleman-Liau values and greater estimated reading difficulty.
Although some overlap between the two distributions was observed, the shift in central tendency provides clear evidence of systematic stylistic differences between the two sources.
The strong and consistent performance across multiple models also provides insight into the role of model complexity. Differences among the strongest models were modest. The ensemble achieved the highest mean AUC, accuracy, and F1 score in the repeated pair-level analysis, but logistic regression, SVM, random forest, and XGBoost performed closely. The nested grouped cross-validation analysis produced the same overall conclusion after hyperparameter tuning. This indicates that the classification task is primarily driven by the quality of the feature representation rather than the complexity of the modeling approach. In other words, once informative features are extracted, even relatively simple models are capable of achieving high performance.
These findings have important implications for the detection of AI-generated misinformation. The results suggest that effective detection systems can be constructed using interpretable, low-dimensional feature sets without requiring highly complex deep learning architectures. This is particularly relevant in practical settings where computational efficiency and model interpretability are important considerations.

The stronger differences observed for readability and lexical features may, in contrast, reflect transformations introduced by the LLM during the rewriting process. Although the model was instructed to preserve the substantive misinformation claim and storyline, it was also asked to modify sentence structure, vocabulary, and organization. Such rewriting may produce more regular sentence construction, smoother transitions, and greater stylistic standardization than are present in the original human-written articles. Accordingly, the prominence of readability, lexical diversity, and related structural features should be interpreted as evidence of detectable stylistic changes associated with the particular LLM rewriting procedure used in this study, rather than as universal characteristics of all AI-generated misinformation.

At the same time, several limitations should be acknowledged when interpreting these findings. Most importantly, the present study examines a controlled paired rewriting setting in which an LLM was provided with an existing human-written misinformation article and instructed to preserve its substantive false claim and overall narrative while modifying its wording, sentence structure, and organization. The classification task is therefore most directly one of distinguishing human-written misinformation from LLM-generated rewrites, rather than detecting AI-generated misinformation in all possible settings. In addition, the AI-generated texts were produced using a single language model and a single prompting framework. The observed linguistic differences and classification performance may therefore depend partly on this particular generation process and may not generalize directly to misinformation generated from scratch, produced using other language models or prompting strategies, or subsequently edited by humans. Future work should evaluate these features across multiple LLM families, prompts, generation paradigms, topics, and levels of human post-editing to determine which signals remain robust across more realistic forms of AI-assisted misinformation.
Additionally, the analysis was based on static text features and did not incorporate contextual or semantic representations. While strong performance was achieved using readability and lexical features, it is possible that more nuanced differences between human and AI-generated content could be captured using advanced language models or embedding-based approaches.


\section*{conclusions}

In this study, the problem of distinguishing human-written fake news from AI-generated fake news was examined using a range of machine learning models and interpretable text-based features. High classification performance was consistently achieved across all models, indicating that meaningful and systematic differences exist between the two types of content.

It was found that readability-based features, particularly the Coleman-Liau index, play a dominant role in classification, while emotion-based features contribute comparatively less. These findings suggest that structural and stylistic properties of text provide stronger signals than emotional tone for identifying AI-generated content. The ensemble achieved the strongest average performance across repeated pair-level splits, although several individual models performed comparably. The persistence of high performance under grouped cross-validation and hyperparameter tuning further supports the robustness of the findings.

Overall, the findings demonstrate that human-written misinformation can be distinguished from LLM-generated rewrites of the same content using relatively simple and interpretable features. As text generation systems continue to evolve, further investigation will be required to assess the robustness of these approaches across different models and generation strategies. Nonetheless, the present study provides evidence that stylistic differences remain a viable and interpretable basis for distinguishing human-written misinformation from LLM-generated rewrites within the controlled setting examined here.




\bibliographystyle{unsrtnat}
\bibliography{references}

\section*{about the student author}
Samuel Jaeger is currently attending Kennesaw State University. He is expected to graduate in the spring of 2029 with a Bachelor of Science in Data Science and Analytics. As a member of Kennesaw State's Undergraduate Research Program, he was paired with a mentor who would help guide him through the completion of this research project. That mentor happened to be Dr. Dhrubajyoti Ghosh, who is a co-author of this research.

\section*{press summary}
This study examines whether human-written misinformation can be distinguished from AI-generated rewrites of the same underlying content using measurable text features. By analyzing readability, vocabulary patterns, sentence structure, and emotional tone, the study found that machine learning models can distinguish the two sources with high accuracy. Readability-related features, especially the Coleman-Liau index, were the strongest indicators, 
with AI-generated rewrites showing greater estimated reading difficulty than the human-written originals.
The findings demonstrate that large language model rewriting can introduce detectable stylistic and structural changes, while also highlighting the need for future research across different models, prompts, and forms of AI-generated misinformation.

\end{document}